\documentclass[letterpaper, 10 pt, conference]{ieeeconf}  

\IEEEoverridecommandlockouts                              

\usepackage{cite}
\usepackage{amsmath,amssymb,amsfonts}
\usepackage{algorithmic}
\usepackage{graphicx}
\usepackage{textcomp}
\usepackage{lipsum}
\usepackage{xcolor}
\usepackage[section]{placeins} 

\title{\LARGE \bf
A Robust Aerial Gripper for Passive Grasping and Impulsive Release using Scotch Yoke Mechanism
}

\author{Rajashekhar V S$^{1}$, Vibha M R$^{1}$, Kaushik Das$^{2}$ and Debasish Ghose$^{1}$
\thanks{$^{1}$Rajashekhar V S, Vibha M R and Debasish Ghose are with Department of Aerospace Engineering, Indian Institute of Science, Bangalore, India
        {\tt\small vsrajashekhar@gmail.com, vibhamr2596@gmail.com, dghose@iisc.ac.in}}%
\thanks{$^{2}$Kaushik Das is with Tata Consultancy Services, Innovations Lab, Bangalore, India 
        {\tt\small kaushik.da@tcs.com}}%
}

\begin{document}

\maketitle
\thispagestyle{empty}
\pagestyle{empty}

\begin{abstract}
Aerial transportation requires a simple yet reliable gripper for picking and placing objects of interest. In this work, we design an aerial gripper for passive grasping and impulsive release of ferrous coated objects. Permanent magnets are used for passive grasping and the Scotch Yoke mechanism is used for providing impulsive force to drop the object. The load carrying capacity of the gripper is calculated theoretically and experimentally. The parameters such as the radius of the rotating disk and length of the slider in the Scotch Yoke mechanism were optimized using weighted geometric programming. The dimensions of the gripper mount were derived considering the various components of the gripper. The gripper was mounted on an Unmanned Aerial Vehicle (UAV) and the tests were done by carrying ferrous coated cuboid shaped objects of different sizes and masses. These tests were done in manual and autonomous mode in the outdoor environment.
\end{abstract}

\section{Introduction}
Unmanned Aerial Vehicles (UAV) are being used in recent times for object transportation \cite{valavanis2015handbook}. While accomplishing this, aerial grippers and manipulators are needed for grasping and carrying objects of interest. Usually, gippers are used for grasping the target objects without manipulating them. The manipulators with grippers as end-effectors are used to position and orient the target objects. Various types of aerial grippers have been made in the past including suction grippers \cite{kessens2016versatile}, claw grippers \cite{pounds2011practical}, magnetic grippers \cite{fiaz2017passive}, compliant grippers \cite{ghadiok2012design} and electro-permanent magnetic grippers (EPM) \cite{bahnemann2017decentralized}. Among these grippers, magnetic and EPM can be efficiently used for picking up ferrous coated objects. 

Most of the grippers found in the literature have one degree of freedom. This is because the 1-DoF aerial grippers are used in scenarios where the vision system in the UAVs can be relied on for perfect positioning accuracy \cite{danko2015parallel}. A recent survey on aerial manipulation \cite{xilun2019review} shows that these type of aerial grippers are commonly used due to ease in modelling, control and fabrication. They are also relatively inexpensive when compared to other kinds of grippers used for aerial grasping.

\begin{figure}[t!] 
\centering
    \includegraphics[scale=0.04]{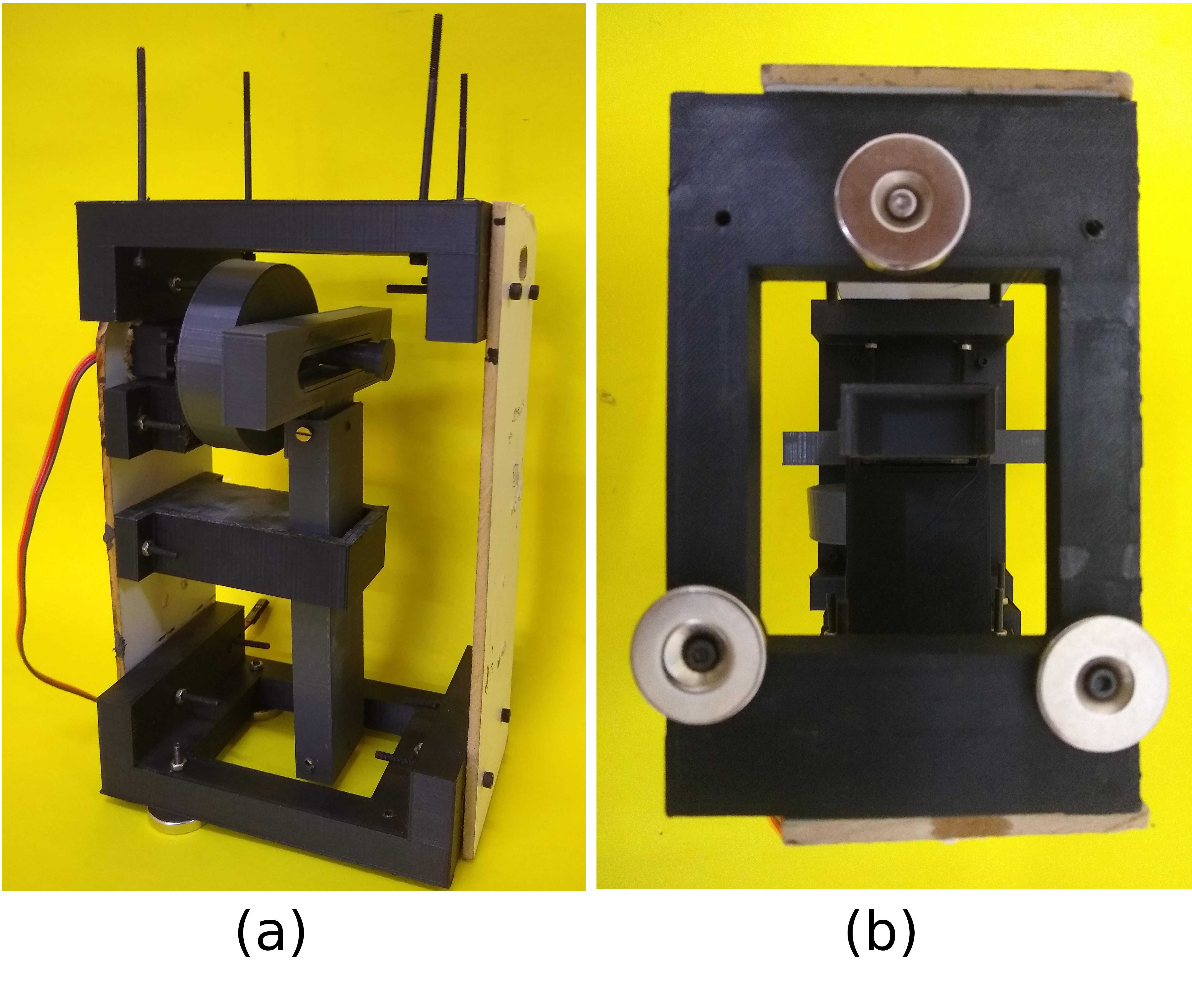} 
\caption{(a) The aerial gripper with the Scotch-Yoke Mechanism (b) The base of the aerial gripper where there are three permanent magnets}
\label{fig_brickgripper}
\end{figure}
There are two methods of grasping and releasing of objects using magnetic grippers, namely, active grasping and active release, and passive grasping and active release. In active grasping and active release, power is required to actuate the electromagnets or electo permanet magnets in order to capture the target object. It is then released by turning off the power supply to the electomagnets. This can be seen in the works of \cite{gawel2017aerial} for aerial picking and delivery. In passive grasping and active release, permanent magnets are used for grasping ferrous coated objects, and an impulsive force is given to detach the object from the UAV. Passive grasping of objects is energy efficient and increases the endurance of the UAV when compared to active grasping using electro-permanent magnets (EPM). This has been experimentally proven in the works of \cite{fiaz2018intelligent}. 

In this work, passive grasping and active release of target objects is done. The gripper has permanet magnets at the bottom for passing grasping and a Scotch Yoke mechanism for active release by providing an impulsive force on the target object. The other contribution of the work focuses on the robust design methodology adopted for optimizing the various dimensions of the aerial gripper.
\section{Motivation and Background}
The design for the gripper was done in order to meet the demands of Challenge 2 of Mohamed Bin Zayed International Robotics Challenge (MBZIRC) - 2020, which requires the Unmanned Aerial Vehicles (UAV) and Unmanned Ground Vehicles (UGV) to build a wall using bricks of different sizes and mass as shown in Figure \ref{fig_brickgripper}(a). In this process, it is necessary to make a gripper for the UAV that is light in weight and consumes low power. The aim is to increase the endurance of the UAV during the challenge. Grippers of this kind can be used for aerial transportation too.

Electro Permanent Magnets (EPM) can be used for picking up ferrous coated objects but they consume more power and are not reliable \cite{fiaz2018intelligent}. Moreover, these EPMs need continuous power supply to produce the required magnetic field \cite{gawel2017aerial}. It was observed during experiments that there was residual magnetism in EPM which caused a delay in demagnetizing and dropping the object. Hence, there is a need for instantaneous grasping and releasing of the payload during operation. Therefore, a gripper with permanent magnets for grasping and Scotch Yoke mechanism for impulsive release has been designed in our work as shown in Figure \ref{fig_brickgripper}(b).
\section{The System Design}
\subsection{Permanent Magnets}
The magnets used for the experiments were ring-shaped neodymium magnets. This shape was chosen because it would aid in easy mounting of the magnets to the base of the gripper mount using bolts and nuts. The neodymium magnet used were of grade N35 and supposed to have a pulling force of 4.66 kg (calculated in Section \ref{subsec_mfd}). But based on our experiments on the ferrous surface, it was found that the pulling force was about 2.1 kg. The ring has an outer diameter of 25 mm and an inner diameter of 5 mm. It has a thickness of 5 mm. The M5 bolts and nuts were used to fix the magnets to the base of the gripper mount. Based on the maximum load to be carried (2 kg), it is found that the repelling forces between the three magnets is minimum when placed 70 mm apart. These three magnets are placed in the corners of an equilateral triangle (of side 70 mm) so that it can pick up the targets objects efficiently.   
\begin{figure}[t!] 
\centering
\includegraphics[scale=0.35]{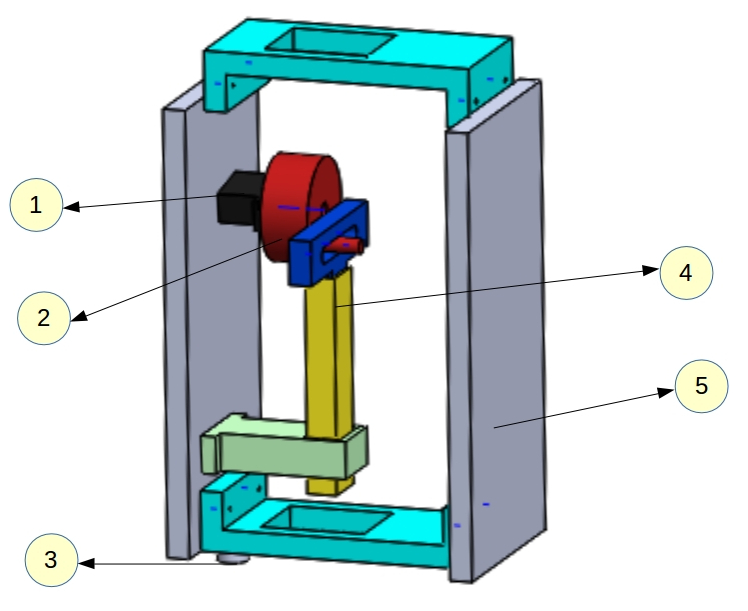} 
\caption{The Computer Aided Design of the Gripper (1) Rotatary actuator (2) Rotating disk (3) Permanent magnet (4) Slider (5) Gripper Mount }
\label{fig_iso}
\end{figure} 
\subsection{Scotch Yoke Mechanism}
The Scotch Yoke mechanism is used because, the actuator can be placed far away from the permanent magnets and also has only two moving links. The Scotch Yoke mechanism \cite{de1945scotch} is used for impulsive release of the target object once it has reached the dropping point. This mechanism has a long life and is highly efficient because the power loss due to friction is eliminated and the wear between the moving parts is minimized. In the gripper, this mechanism is actuated using a servo motor. One of the main reason for chosing this mechanism is that the length of the mechanism is scalable without compromising the torque. This means that the rotary motion of the disk can be transferred to linear motion and this linear motion can be transmitted to a longer distance. 
\subsection{Gripper Mount}
\label{subsec_mount}
The mount for the gripper is the critical part as it has to hold the scotch yoke mechanism and carry the load of the target object. The top portion of the mount has to be fitted to the base of the UAV. The side of the gripper mount has the mechanism fixed to it. The base has three ring shaped neodymium magnets attached to it. There is a slot at the bottom through which the slider of the scotch yoke mechanism comes out and plunges on the target object attached to the magnets by giving an impulsive force.   
\subsection{Unmanned Aerial Vehicle}
The gripper was attached to a UAV of dimensions 1668 mm x 1518 mm x 759 mm and had a mass of 9.6 kg (including the batteries). It has a maximum take off weight of 15.1 kg and a hovering time of 16 minutes. There were six batteries that were used to power the UAV. The total mass of the gripper was 0.534 kg and was mounted close to the center of gravity of the UAV.
\subsection{Flight Control System}
The \textit{NVIDIA Jetson TX2} is used to implement upper level control of position in the UAV. The signals used in the feedback loop are position values from the inertial measurement unit (lateral and longitudinal position from the IMU and the vertical position from the LIDAR sensor). A camera is used for vision based tracking. This controller gives velocity commands to the UAV for the desired navigation. The Robot Operating System (ROS) was used to implement the control algorithm. 
\subsection{Robust Gripper Design}
\begin{figure*}[h] 
\centering
\vspace{0.5cm}
\includegraphics[scale=0.23]{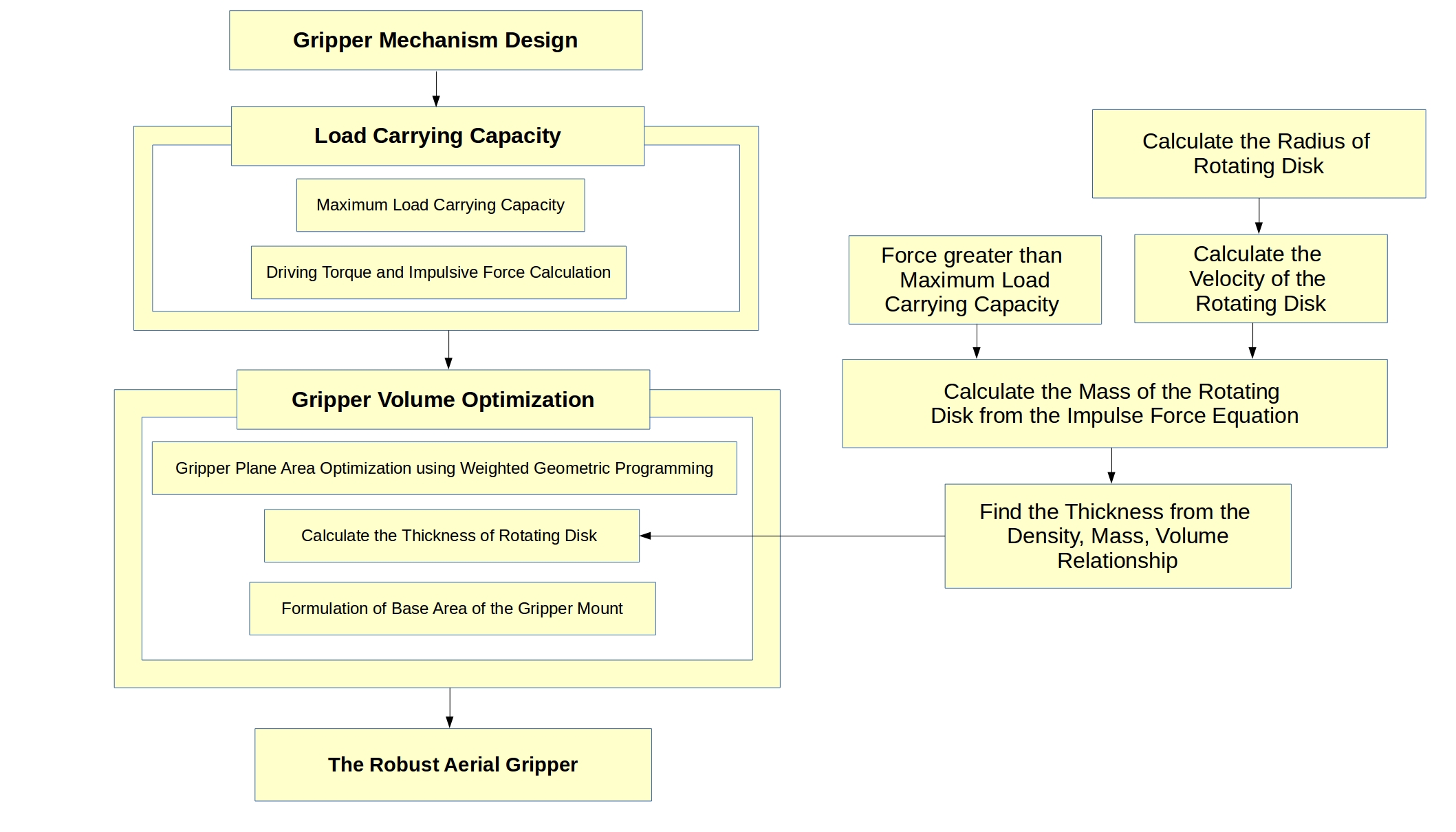} 
\caption{The robust design stages}
\label{fig_brief}
\end{figure*}

The method followed to make the gripper robust is shown in Figure \ref{fig_brief}. The aerial gripper is made robust by calculating the load carrying capacity and optimizing the volume. The load carrying capacity is established by finding the maximum load carrying capacity of the permanent magnets attached below the drone. Then the driving torque required to actuate the mechanism is calculated. The impulsive force required to release the object is determined based on the maximum load carrying capacity. The dimensions of the rotating disk and the slider are optimized using weighted geometric programming. The minimum mass of the rotating disk of the mechanism is calculated from the impulse force equation. From this mass, using the density, mass and volume relation, the thickness of the rotating disk is calculated. Then the base area of the gripper mount is formulated depending on the dimensions of various components of the gripper. Through this procedure, a robust aerial gripper with optimal volume for a given load carrying capacity has been designed.   
\section{Design of the Gripper Mechanism}
\label{sec_lcc}
\subsection{Maximum Load Carrying Capacity}
\label{subsec_mfd}
The maximum load carried is calculated using the relation,
\begin{equation}
\label{equ_loadcarry}
m_{\max}=\frac{B^{2}A}{2g\mu_{0}} 
\end{equation}
where, $B$ is magnetic flux density (T), $A$ is surface area of magnet (m$^2$), $g$ is the acceleration due to gravity (m/s$^2$), $\mu_{0}$ is permeability of air (Tm/A).

In this case, $B$=0.494 T, $A$=0.000471239 m$^2$, $g$=9.81 m/s$^2$ and $\mu_{0}$=1.25663753$\times$10$^{-6}$ Tm/A. Substituting these values in Equation \ref{equ_loadcarry}, we get $m_{max}$=4.66 kg. Experimental results showed that each magnet was capable of lifting a maximum mass of $m_{max}$=2.1 kg. This could be due to the variation in the magnetic flux density and the material properties.

To establish static equilibrium of the object after grasp, three points of contacts are needed. Hence three magnets are used for grasping. Moreover, the maximum load it would carry is approximately 6 kg. This can carry a 2 kg payload under low wind conditions which is our requirement.  
\subsection{Driving Torque and Impulsive Force}
\begin{figure}[h!] 
\centering
\includegraphics[scale=0.3]{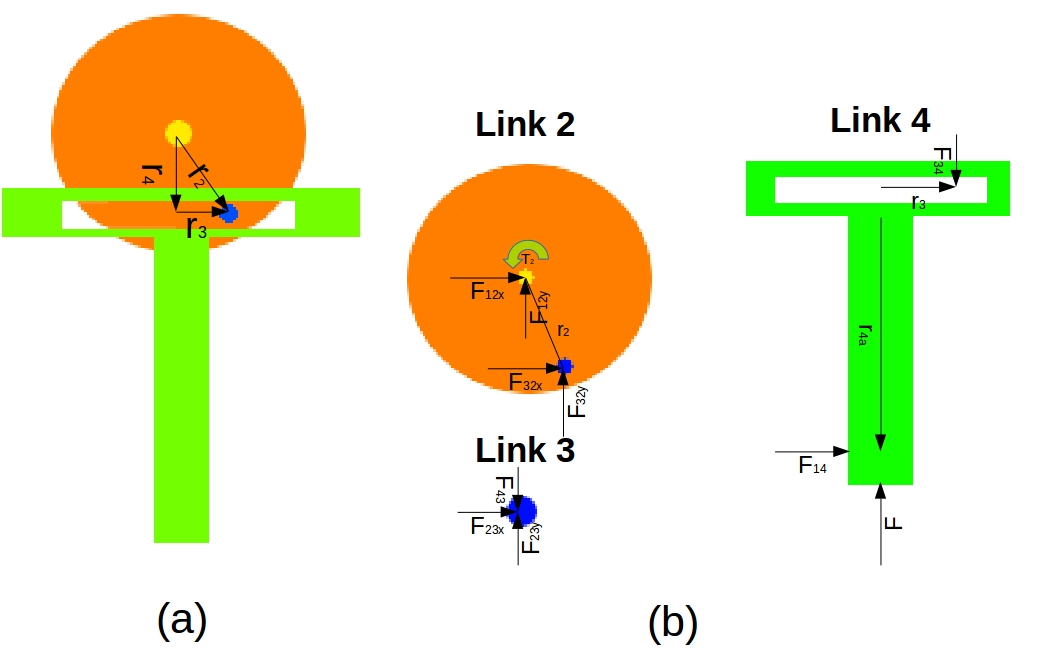} 
\caption{(a) The vector loop diagram used for the kinematics of the mechanism (b) The forces and torques acting on different links of the mechanism}
\label{fig_dynkin}
\end{figure}
\subsubsection{Kinematics of the Mechanism}
The loop closure equation is as follows from Figure \ref{fig_dynkin} (a).
\begin{equation}
r_{2}=r_{3}+r_{4}
\end{equation}
It can be written as,
\begin{equation}
\label{equ_loop}
\begin{split}
r_{2}\cos \theta_{2} &= r_{3} \cos \theta_{3}+r_{4} \cos \theta_{4}\\
r_{2}\sin \theta_{2} &= r_{3} \sin \theta_{3}+r_{4} \sin \theta_{4}
\end{split}
\end{equation}
It can be seen from the Figure \ref{fig_dynkin} that the value of $\theta_{2}$ is the input angle and $r_{2}$ is known, $\theta_{3}=0$ and $\theta_{4}=-\pi/2$. Therefore, substituting them in Equation \ref{equ_loop}, we get 
\begin{equation}
\begin{split}
r_{4} &=-r_{2} \sin \theta_{2}\\
r_{3} &=r_{2} \cos \theta_{2}
\end{split}
\end{equation}
The velocity equations are as follows.
\begin{equation}
\label{equ_velocity}
\begin{split}
\dot{r_{4}} &=-r_{2} \cos \theta_{2} \dot{\theta_{2}}\\
\dot{r_{3}} &=-r_{2} \sin \theta_{2} \dot{\theta_{2}}
\end{split}
\end{equation}
The acceleration equations are as follows.
\begin{equation}
\begin{split}
\ddot{r_{4}} &=r_{2} \sin \theta_{2} \ddot{\theta_{2}}\\
\ddot{r_{3}} &=-r_{2} \cos \theta_{2} \ddot{\theta_{2}}
\end{split}
\end{equation}
\subsubsection{Dynamics of the Mechanism}
Based on the Newton-Euler method, the force and momentum equations are as follows from Figure\ref{fig_dynkin} (b).\\
Link 4: \\
\begin{equation}
\label{equ_l4}
\begin{split}
-F+F_{34} &=0\\
F_{14}r_{4a}-F_{34}r_{3} &=0
\end{split}
\end{equation}
Link 3: \\
\begin{equation}
\begin{split}
F_{23x} &=0\\
F_{34}-F_{23y} &=0
\end{split}
\end{equation}
Link 2: \\
\begin{equation}
\label{equ_l2}
\begin{split}
F_{12x}-F_{23x} &=0\\
F_{12y}-F_{23y} &=0\\
T_{2}+r_{2} \cos \theta_{2}F_{23x}+ r_{2} \sin \theta_{2}F_{23y} &=0
\end{split}
\end{equation}
The Equations \ref{equ_l4} to \ref{equ_l2} can be converted into the form as follows.
\[
\begin{bmatrix}
1 & 0 & 0 & 0 & 0 & 0 & 0 \\
-r_{2} & r_{4a} & 0 & 0 & 0 & 0 & 0 \\
 0 & 0 & 1 & 0 & 0 & 0 & 0 \\
  1 & 0 & 0 & -1 & 0 & 0 & 0 \\
    0 & 0 & -1 & 0 & 1 & 0 & 0 \\
   0 & 0 & 0 & -1 & 0 & 1 & 0 \\
   0 & 0 & r_{2}\cos \theta_{2} & r_{2}\sin \theta_{2} & 0 & 0 & 1 \\
\end{bmatrix}
\begin{bmatrix}
F_{34} \\ F_{14} \\  F_{23x} \\  F_{23y} \\  F_{12x} \\  F_{12y} \\  T_{2}
\end{bmatrix}
=
\begin{bmatrix}
F \\ 0 \\ 0 \\ 0 \\ 0 \\ 0 \\ 0
\end{bmatrix}
\] 
Solving this we get, 
\begin{equation}
\label{equ_dynfinal}
\begin{bmatrix}
F_{34}\\ 
F_{14}\\ 
F_{23x}\\ 
F_{23y}\\ 
F_{12x}\\ 
F_{12y}\\ 
T_{2}
\end{bmatrix}
=
\begin{bmatrix}
F\\ 
\frac{r_{3}F}{r_{4a}}\\ 
0\\ 
F_{23y}\\ 
0\\ 
F\\ 
-r_{2}\sin \theta_{2} F
\end{bmatrix}
\end{equation} 
It can be seen in Equation \ref{equ_dynfinal} that there is no force acting on the X-axis of the plane. The torque ($T_{2}$) equation found here will be used for optimizing the gripper volume in the next section. 
\subsubsection{Impulsive Force}
The impulse equation is as follows:
\begin{equation}
I=p_{f}-p_{i}=mv_{f}-mv_{i}
\end{equation}
The average force exerted by the object on slider of the mechanism is,
\begin{equation}
\label{equ_impulsiveforce}
F_{avg}=\frac{mv_{f}-mv_{i}}{t_{f}-t_{i}}
\end{equation}
\section{Optimization of Gripper Volume}
\subsection{Gripper Plane Area Optimization}
\label{subsec:gpgpareaopti}
The Scotch Yoke is a planar mechanism and hence the area occupied by the parts play a role in deciding the height and depth of the gripper mount. Here the height by depth area is minimized and the torque produced by the actuator is maximized. The only constraint is that, the ratio between the radius of the rotating disk and the height of the slider should be less than $0.25$ due to the space constraint below the drone. Hence, it is a multi-objective optimization problem with posynomials in the objective function and a monomial in the constraint. Therefore, it is solved by using weighted geometric programming \cite{ojha2015hybrid}. \\
The objective function is \\
\begin{equation}
\begin{split}
\min \; f(r_{2}, r_{4a}) &=4r_{2}^{2}+2r_{2}r_{4a}\\
\max \; f(r_{2}) &= 75r_{2}\\
\end{split}
\end{equation}
subject to
\begin{equation}
\begin{split}
r_{2}r_{4a}^{-1}\leq 0.25\\
r_{2},r_{4a}\geq \epsilon
\end{split}
\end{equation}
The above can be rewritten as,
\begin{equation}
\begin{split}
\min \; f(r_{2}, r_{4a})=4r_{2}^{2}+2r_{2}r_{4a}\\
\min \; f(r_{2})= 0.013333 r_{2}^{-1}\\
\end{split}
\end{equation}
subject to
\begin{equation}
\begin{split}
4r_{2}r_{4a}^{-1}\leq 1\\
r_{2},r_{4a}\geq \epsilon
\end{split}
\end{equation}

Since it is a multi-objective optimization, weights are assigned to the objective function such that $w_{1}+w_{2}=1, w_{1},w_{2}>0$. This problem has a degree of difficulty one. The dual program is:
\begin{equation}
\begin{split}
\max \; V(w)=\left(\frac{4w_{1}}{\lambda_{1}}^{\lambda_{1}}\right)\left(\frac{2w_{1}}{\lambda_{2}}^{\lambda_{2}}\right)\left(\frac{0.013333w_{2}}{\lambda_{3}}^{\lambda_{3}}\right)\\ \left(\frac{4}{\lambda_{4}}^{\lambda_{4}}\right)\left(\lambda_{4}^{\lambda_{4}}\right)
\end{split}
\end{equation}
subject to\\
\begin{equation}
\label{equ_gpdual2}
\begin{split}
\lambda_{1}+\lambda_{2}+\lambda_{3}=1\\
2\lambda_{1}+\lambda_{2}-\lambda_{3}+\lambda_{4}=0\\
\lambda_{2}-\lambda_{4}=0\\
w_{1}+w_{2}=1\\
w_{1},w_{2}>0\\
\lambda_{1},\lambda_{2},\lambda_{3},\lambda_{4}>0
\end{split}
\end{equation}

Solving the conditions in Equation \ref{equ_gpdual2}, we get $\lambda_{1}=\frac{1}{6}$, $\lambda_{2}=\frac{1}{6}$, $\lambda_{3}=\frac{2}{3}$, $\lambda_{4}=\frac{1}{6}$. By considering different weights, the maximum value of dual objective function is found. From them, the values of $r_{2}$ and $r_{4a}$ are obtained and given in Table \ref{table_gp}.
\begin{table}[h!]
\begin{center}
\caption{Solutions of the dual relationship}
\label{table_gp}
\begin{tabular}{|c|c|c|c|c|}
\hline
$w_{1}$  & $w_{2}$  & $r_{2}$           & $r_{4a}$            & V(w)         \\ \hline
0.1 & 0.9 & 0.0655802278 & 0.2623209111 & 0.1032183906 \\ \hline
0.2 & 0.8 & 0.07077726   & 0.2831090401 & 0.1202260929 \\ \hline
0.3 & 0.7 & 0.0724288584 & 0.2897154336 & 0.1259025487 \\ \hline
0.4 & 0.6 & 0.0721803804 & 0.2887215215 & 0.1250401754 \\ \hline
0.5 & 0.5 & 0.0704980471 & 0.2819921883 & 0.1192793914 \\ \hline
0.6 & 0.4 & 0.0674637726 & 0.2698550902 & 0.1092326546 \\ \hline
0.7 & 0.3 & 0.0628900842 & 0.2515603366 & 0.0949239044 \\ \hline
0.8 & 0.2 & 0.0561759485 & 0.224703794  & 0.0757376926 \\ \hline
0.9 & 0.1 & 0.0454707903 & 0.1818831612 & 0.0496222265 \\ \hline
\end{tabular}
\end{center}
\end{table}

In our case, more weightage should be given for minimizing the area than maximizing the torque. From Table \ref{table_gp} it can be observed that the minimum $r_{2}$ and $r_{4a}$ are obtained when the values of $w_{1}=0.9$ and $w_{2}=0.1$. Thus the optimal solution of $r_{2}=0.0454707903$ m and $r_{4a}=0.1818831612$ m. 
\subsection{Thickness of the Rotating Disk}
\label{subsec:thickness}
The thickness of the gripper is calculated using the procedure shown in the flowchart in Figure \ref{fig_brief}. The value of $r_{2}$ calculated in Section \ref{subsec:gpgpareaopti} is substituted in  $\dot{r_{4}}$ of Equation \ref{equ_velocity}. The average force exerted by the object on the slider should be atleast 1.2 times greater than the maximum load carried in order to detach the load due to impulse. This is found to be about $75$N. These values are substituted in Equation \ref{equ_impulsiveforce} and the unknown mass value is found. Since the density ($\rho$) of the material used, the mass ($m$), the value of $r_{2}$ are known and the thickness of the disk is unknown, it is found by the relation Density=$\frac{Mass}{Volume}$. Similar procedure can be followed to obtain the thickness of the slider.    
\subsection{Gripper Mount Dimensions Formulation}
\label{subsec:grippermountdim}
The length of the gripper mount ($L_{gm}$), the height of the gripper mount ($H_{gm}$), the depth of the gripper mount ($D_{gm}$) are given by 
\begin{equation}
\begin{split}
L_{gm}=2r_{2}+2m_{d}\\
H_{gm} =2r_{2}+r_{4a}\\
D_{gm}=A_{l}+D_{t}+S_{t}+2m_{d}
\end{split}
\end{equation}
where, $r_{2}$ is the radius of the rotating disk, $m_{d}$ is the diameter of the permanent magnet, $r_{4a}$ is the length of slider $A_{l}$ is the length of the actuator, $D_{t}$ is the thickness of the rotating disk and $S_{t}$ is the thickness of the slider. 
\subsection{Dimensions of the Robust Aerial Gripper}
A factor of safety of $2$ has been chosen for the radius of the rotating disk since it is the crucial part of the mechanism as it transfers the torque from the rotary actuator to the slider. The length of the slot in the slider is $2r_{3}$. The length of the slider $r_{4a}$ is calculated in Section \ref{subsec:gpgpareaopti}. The thickness of the disk ($D_{t}$) and the slider ($S_{t}$) are found using the method explained in Section \ref{subsec:thickness}. The dimensions of the gripper mount such as length ($L_{gm}$), depth ($D_{gm}$) and height ($H_{gm}$) are calculated from Section \ref{subsec:grippermountdim}. The values of these parameters used are listed in Table \ref{table_dimparts}.  
\begin{table}[h!]
\begin{center}
\caption{The dimensions of various parts of the aerial gripper}
\label{table_dimparts}
\begin{tabular}{|c|c|}
\hline
Components                            & Dimension (cm) \\ \hline
Radius of the rotating disk ($r_{2}$)      & 2.5            \\ \hline
Length of the slot in the slider      & 5              \\ \hline
Length of the slider ($r_{4a}$)      & 18             \\ \hline
Length of the actuator ($A_{l}$)           & 3              \\ \hline
Thickness of the disk ($D_{t}$)            & 2.5            \\ \hline
Thickness of the slider ($S_{t}$)          & 2.5            \\ \hline
Diameter of the permanent magnet ($m_d$) & 2.5            \\ \hline
Length of the gripper mount ($L_{gm}$)     & 10             \\ \hline
Depth of the gripper mount ($D_{gm}$)      & 13             \\ \hline
Height of the gripper mount ($H_{gm}$)      & 23             \\ \hline
\end{tabular}
\end{center}
\end{table}
\section{Experimental Results}
The fabrication of the gripper was done using the dimensions in the Table \ref{table_dimparts}. The brick lifting experiments were initially carried out with the gripper before mounting it on the UAV. The bricks were found to be intact with the gripper at various orientations after gripping.
\begin{figure*}[h]
\centering
\vspace{0.5cm}
\includegraphics[scale=0.08]{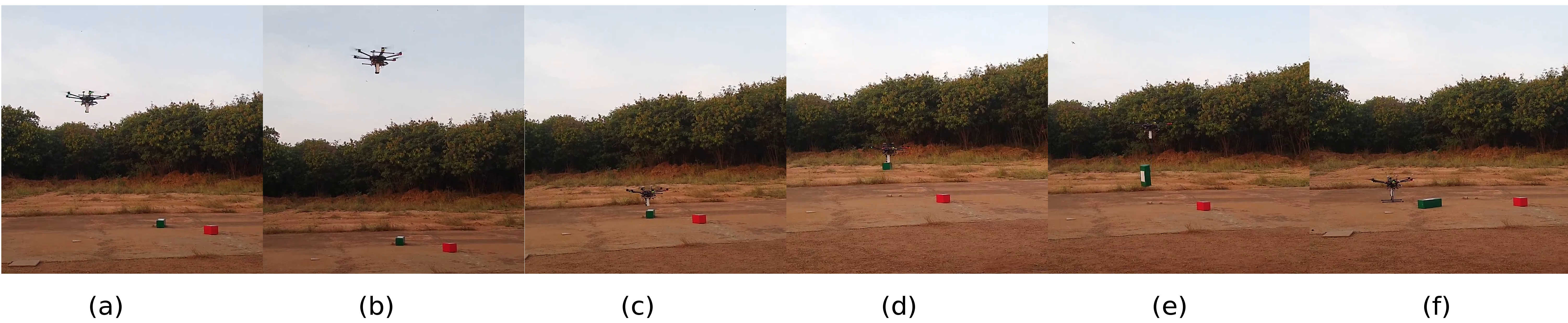} 
\caption{Bricks being lifted (a) Autonomous take-off of the UAV with the Aerial Gripper (b) Visual servoing (c) Approaching the target object (d) Picking up the target object (e) Dropping the target object (f) Autonomous landing of the the UAV }
\label{fig_auto}
\end{figure*} 
\subsection{Fabrication of the Aerial Gripper}
The parts of the aerial gripper were modelled in a Computer Aided Design (CAD) software. These parts were then 3D printed and assembled. The three magnets were attached below the gripper mount. The sides of the gripper mount were made of hylam sheet. A servo motor with a torque capacity of $20$ kg-cm at $6$ V was used to actuate the mechanism. 
\subsection{Working of the Aerial Gripper}
\begin{figure}[h!] 
\centering
\includegraphics[scale=0.045]{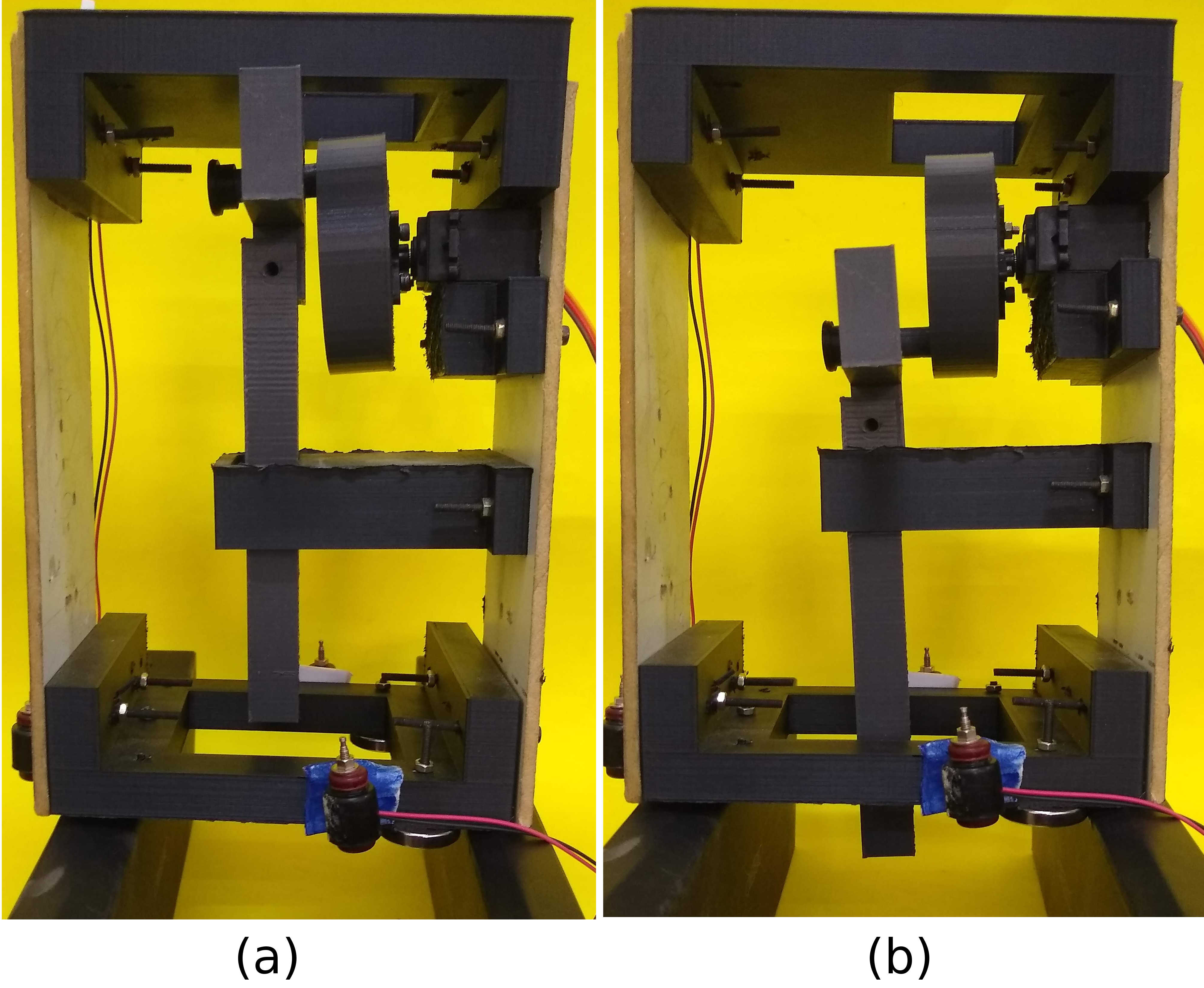} 
\caption{(a) The slider is at the top. The brick can be attached to the gripper. (b) The slider is at the bottom. The brick would be detached at this state}
\label{fig_working}
\end{figure}

There are two possible states of the gripper. In Figure \ref{fig_working}(a), the slider is at top and the brick can be attached to the gripper. The UAV can then transport the brick to the desired location. The servo motor is then actuated, and the slider moves down and pushes the brick with an impulse. Due to this impulse, the brick is detached from the UAV. The position of the slider at the bottom end is shown in Figure \ref{fig_working}(b). Once the brick is dropped, the slider goes back to the initial position.      
\subsection{Experimental Conditions}
\begin{figure}[h!] 
\centering
\includegraphics[scale=0.07]{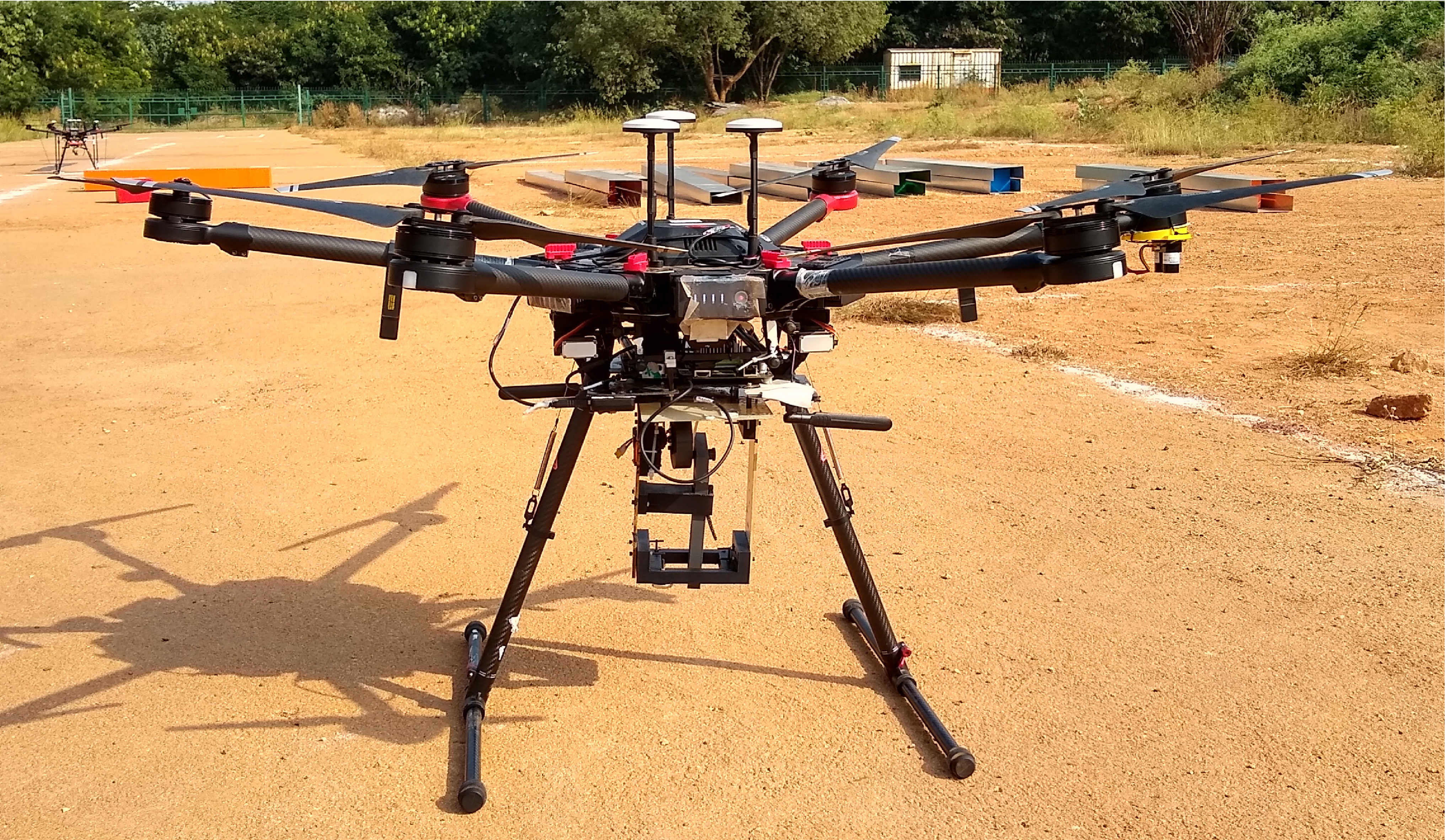} 
\caption{The aerial gripper is mounted below the UAV.}
\label{fig_expdrone}
\end{figure}

The experiments were carried out in the outdoor environment by placing the bricks of various sizes and shapes in the channel. This was done to prevent the bricks from getting displaced due to downwash effect of the UAV. The aerial gripper was mounted below the UAV using the mount designed in Section \ref{subsec_mount}. It is shown in Figure \ref{fig_expdrone}. The signal to the servo was given through the Arduino Nano Microcontroller which receives commands from the Jetson TX2 board mounted on the drone.   
\subsection{Objects Picked}
\begin{figure}[h!] 
\centering
\includegraphics[scale=0.14]{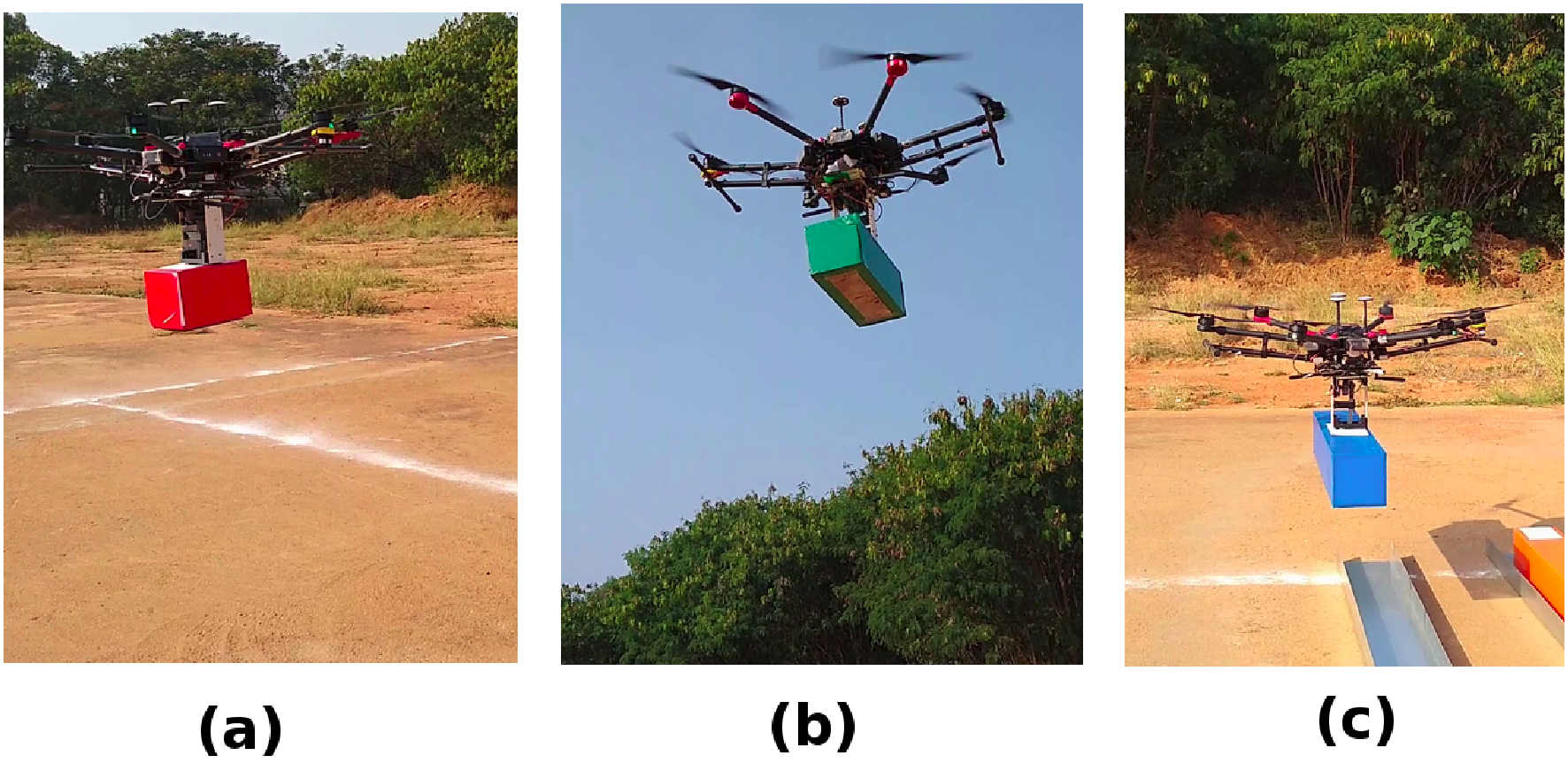} 
\caption{Bricks being lifted (a) The red brick of mass 1kg (b) The green brick of mass 1 kg (c) The blue brick of mass 1.5 kg}
\label{fig_expresults}
\end{figure}

The bricks were picked from the channel and placed in the dropping zone. The efficacy of the aerial gripper was tested by picking the red brick of mass 1kg as shown in Figure \ref{fig_expresults}(a), green brick of mass 1kg as shown in Figure \ref{fig_expresults}(b) and blue brick of mass 1.5kg as shown in Figure \ref{fig_expresults}(c). These bricks were dropped once they reached the desired dropping point by actuating the Scotch Yoke mechanism. 
\subsection{Experiments in Autonomous Mode}
In the autonomous mode, the UAV took off autonomously with the aerial gripper. Then visual servoing was done to locate the ferrous coated brick. Once the brick was identified, the UAV approached the brick, picked it up and went to a certain height. Then, it dropped the brick and landed autonomously. The snapshots of the experiments can be seen in Figure \ref{fig_auto} (a-f). The entire process was completed in less than $65$ seconds.  
\section{Conclusions}
In this work, we have designed an aerial gripper that can be used for passive grasping and impulsive release of objects. The load carrying capacity of the gripper was calculated and the driving torque required for the Scotch Yoke mechanism was derived. The various lengths of the gripper were optimized. Finally experiments were done by mounting the gripper on the UAV in manual and autonomous mode. The future works lie in designing the base of the gripper in such a way that it could pick up objects of different shapes with feedback.

\textbf{Acknowledgment}
The authors wish to acknowledge partial funding support from RBCCPS/IISc and Khalifa University, Abu Dhabi. The authors also thank the IISc-TCS MBZIRC team for their help in conducting the experiments.

\bibliographystyle{IEEEtran}
\bibliography{mybibfile}

\end{document}